\documentclass[10pt,conference]{IEEEtran}
\IEEEoverridecommandlockouts
\usepackage{cite}
\usepackage{amsmath,amssymb,amsfonts}
\usepackage{algorithmic}
\usepackage{graphicx}
\usepackage{textcomp}
\usepackage{xcolor}
\usepackage{multirow}
\usepackage{capt-of}    
\def\BibTeX{{\rm B\kern-.05em{\sc i\kern-.025em b}\kern-.08em
    T\kern-.1667em\lower.7ex\hbox{E}\kern-.125emX}}

\DeclareMathOperator{\qa}{\mathcal{QA}}    

\usepackage{url}
\usepackage{listings}
\usepackage{comment}
\usepackage{xspace}
\usepackage{hyperref}
\usepackage{multirow}
\usepackage{graphicx}

\begin{document}

\title{\acronym: An Exemplar for\\ Self-Adaptive Underwater Vehicles
\thanks{This work was supported by the European Union’s Horizon 2020 Framework Programme through the MSCA network REMARO (Grant Agreement No 956200).}
}

\newcommand\red[1]{{\color{red}#1}}
\newcommand\blue[1]{{\color{blue}#1}}
\newcommand\acronym{SUAVE\xspace}
\newcommand\kw[1]{\texttt{{\small{#1}}}}

\author{%
	\IEEEauthorblockN{%
		Gustavo Rezende Silva\IEEEauthorrefmark{1},
        Juliane P{\"a}{\ss}ler\IEEEauthorrefmark{2},
        Jeroen Zwanepol\IEEEauthorrefmark{1},
        Elvin Alberts\IEEEauthorrefmark{3}\IEEEauthorrefmark{1}, 
        S. Lizeth Tapia Tarifa\IEEEauthorrefmark{2},\\
	    Ilias Gerostathopoulos\IEEEauthorrefmark{3},        
		Einar Broch Johnsen\IEEEauthorrefmark{2},
            Carlos Hern{\'a}ndez Corbato\IEEEauthorrefmark{1}
	}%
	\IEEEauthorblockA{\IEEEauthorrefmark{1}\textit{Technical University of Delft, Delft, The Netherlands}\\Email: \{g.rezendesilva,c.h.corbato\}@tudelft.nl}{j.m.zwanepol@student.tudelft.nl}%
    \IEEEauthorblockA{\IEEEauthorrefmark{2}\textit{University of Oslo, Oslo, Norway}\\Email: \{julipas,sltarifa,einarj\}@ifi.uio.no}%
	\IEEEauthorblockA{\IEEEauthorrefmark{3}\textit{Vrije Universiteit Amsterdam, Amsterdam, The Netherlands}\\Email: \{e.g.alberts,i.g.gerostathopoulos\}@vu.nl}%
    
}

\maketitle

\begin{abstract}
  Once deployed in the real world, autonomous underwater vehicles
  (AUVs) are out of reach for human supervision yet need to take
  decisions to adapt to unstable and unpredictable environments. To
  facilitate research on self-adaptive AUVs, this paper presents
  \acronym, an exemplar for two-layered system-level adaptation of
  AUVs, which clearly separates the application and self-adaptation
  concerns. The exemplar focuses on a mission for underwater pipeline
  inspection by a single AUV, implemented as a ROS2-based system. This
  mission must be completed while simultaneously accounting for
  uncertainties such as thruster failures and unfavorable
  environmental conditions.  The paper discusses how \acronym can be
  used with different self-adaptation frameworks, illustrated by an
  experiment using the Metacontrol framework to compare AUV behavior
  with and without self-adaptation.  The experiment shows that the use
  of Metacontrol to adapt the AUV during its mission improves its
  performance when measured by the overall time taken to complete the
  mission or the length of the inspected pipeline.
\end{abstract}


\section{Introduction}

Autonomous robots are an excellent case for applying self-adaptation
techniques
\cite{edwardsArchDrivenRobotics,camara_software_2020,park_task-based_2012,shin_platooning_2021,askarpour_robomax_2021,undersea,ACROS}. These
robots face uncertainty in their operation stemming from both the
system (e.g., sensor failures) and the environment (e.g., different
terrains). They need to complete their missions despite such
uncertainty \cite{Ludvigsen-2021} with minimal or no human supervision
\cite{Huang-2004}.  A subclass of these robots, autonomous underwater
vehicles (AUVs)\cite{Wynn-2014} which are used for, e.g., subsea
observation, are particularly challenging: once they have been
deployed in the real world, they need to take both low-level (e.g.,
increase thruster power) and high-level (e.g., dive deeper) adaptive
decisions without \emph{any} human supervision.

Self-adaptive systems can be implemented as two-layered systems
consisting of a \emph{managed} and a \emph{managing}
subsystem~\cite{weyns2020introduction}. The managed subsystem handles
the domain concerns, while the managing subsystem implements the
adaptation logic and exploits functional alternatives of the managed
subsystem to handle the self-adaptation process.

This paper proposes the exemplar \acronym\footnote{ Self-adaptive
  Underwater Autonomous Vehicle Exemplar.} to facilitate research in
the challenging domain of self-adaptive AUVs and to allow the
comparison of different self-adaptation strategies.  \acronym is based
on ROS2 -- one of the most widely adopted robotics software
frameworks~\cite{doi:10.1126/scirobotics.abm6074}. This ensures that
the system built for \acronym can (i)~run directly on real robots and
not only in simulation environments, (ii)~serve as a basis for other
adaptive robotic missions, and (iii)~be easily extended with new
functionalities and adaptation concerns. The exemplar is publicly
available at \url{https://github.com/kas-lab/suave}.

The exemplar focuses on the scenario of \emph{pipeline inspection} for
a single AUV.  The AUV's mission is to first search for a pipeline on
the seabed, then follow and inspect the pipeline. The functionalities
required to accomplish this mission are implemented in the
\emph{managed subsystem} of \acronym.  During the execution, of the
mission, two types of uncertainties are considered: component failures
in the form of thruster failures (e.g., due to debris getting stuck in
a thruster) and changes in the environmental conditions in the form of
changes in the water visibility (e.g., due to currents disturbing
sediment from the seabed). While the first uncertainty may impact the
robot's motion by making it move unexpectedly, the second impacts the
efficiency of the pipeline search and detection by forcing the robot
to be closer to the seabed to detect the pipeline in case of poor
water visibility, which results in a smaller field of view while
searching.

The exemplar enables the development of a \emph{managing subsystem} to
address the previous uncertainties. The managing subsystem should be
able to monitor the current runtime circumstances, recover the AUV's
thrusters in case of a thruster failure, and adjust the AUV's path
generation algorithm to account for changes in water visibility.

To illustrate the use of adaptation frameworks in \acronym, the
managing subsystem was implemented with
Metacontrol~\cite{corbato2013model,bozhinoski22advrob}, a framework
that enables self-adaptation in robotic systems and promotes the reuse
of the adaptation logic by exploiting a model of the managed subsystem
at runtime. Metacontrol's strength lies in the separation between the
application and adaptation concerns, i.e., in the separation between
the robot's operation and the logic of when and how to adapt.  This
separation of concerns allows the adaptation logic to be reused in a
straightforward way in different applications.  However, it is
important to highlight that even though \acronym is equipped with a
Metacontrol-based adaptation logic, the exemplar can also be used
without Metacontrol, which in addition allows for comparing other
approaches to Metacontrol-based ones.

In summary, the contributions of this paper are:
\begin{itemize}
\item a \emph{self-adaptation exemplar for AUVs using ROS2} that can
  be equipped with different adaptation logics, enables the comparison
  of different self-adaptation strategies, forms a basis for other
  adaptive robotic missions, and can run both on real robots and in
  simulation environments; and
\item a \emph{Metacontol-based adaptation logic formulation} that can
  serve as a baseline for future research and as a benchmark for
  self-adaptation strategies, and is easily reusable for other robotic
  and non-robotic applications.
\end{itemize}

\emph{Paper outline.}  Section~\ref{sec:related} presents related
work, after which Section~\ref{sec:problem_statement} further details
the use case and the overall architecture. The managed subsystem is
described in Section~\ref{sec:managed}, while
Section~\ref{sec:managing_subsystem} discusses the managing subsystem
and how Metacontrol is applied to the use
case. Section~\ref{sec:managing_requirements} briefly explains how the
exemplar can be reused and extended, and Section~\ref{sec:results}
presents and discusses the results of applying Metacontrol. Finally,
Section~\ref{sec:conclusion} concludes the paper.

\section{Related work} \label{sec:related}

The UNDERSEA exemplar by Gerasimou \emph{et
al.}~\cite{undersea} provides an AUV simulation in which the robot performs self-adaptation to
deal with uncertainties such as sensor failures and changing goals. 
\acronym is related to UNDERSEA as both address the domain of self-adaptive AUVs. However, a
key difference is the underlying libraries used to develop software
for the robot. UNDERSEA uses MOOS-IvP while \acronym uses
ROS2, a more widely used framework that is considered state of the art in the robotics
research community, which contributes to the reusability and extensibility of
\acronym.

There have been previous exemplars that do use ROS, in particular, the
Body Sensor Network by Gil \emph{et al.}~\cite{bsnexemplarROS}. However,
its application differs significantly from \acronym as it concerns
health monitoring through a series of sensors rather than a robot
vehicle fulfilling a mission autonomously.

Cheng \emph{et al.} proposed AC-ROS~\cite{ACROS}, a framework 
which uses assurance cases to endow a ROS-based system with
self-adaptive capabilities. Specifically, it concerns an `EvoRally'
vehicle, a terrestrial robot tasked with patrolling an environment as
its mission, while meeting requirements such as energy
efficiency. The authors 
do not provide the source code of the proposed system,
which means it does not serve as an exemplar as \acronym
does.

The paper by Bozhinoski \emph{et al.} \cite{bozhinoski22advrob} concerns an earlier iteration of using MROS for runtime adaptation similar to this paper. Their work revolves around two cases, a manipulator robot with a ``pick and place" task and a mobile robot navigating around obstacles on a factory floor. Both of the use cases show a need to deal with uncertainties, e.g., with a safety concern by disabling one of the pick and place arms. When compared to \acronym, the key differences are the migration from ROS to ROS2, as well as the use case being an AUV rather than a manipulator or mobile terrestrial robot. 
\section{Pipeline inspection exemplar}
\label{sec:problem_statement}
This section describes the use case and system architecture, the two
system layers are detailed in Sections~\ref{sec:managed} and
\ref{sec:managing_subsystem}.

\subsection{Use case description}
\label{sec:use_case_description}
The use case in this exemplar is about an AUV inspecting pipelines located on a seabed.
Its mission consists of two sequential tasks, $(T1)$~searching for the pipeline, then $(T2)$~simultaneously following and inspecting the pipeline.

When performing its mission, the AUV is subject to two sources of
uncertainty that could trigger self-adaptation: $(U1)$~thruster
failures and $(U2)$~changes in water visibility. $U1$~arises from the
possibility of the AUV's thrusters failing at runtime, which may cause
the AUV to move unexpectedly. This is relevant for both $T1$ and $T2$. To overcome $U1$, the managed subsystem of the AUV
contains functional alternatives. When
one or more thrusters fail, it is possible to enter a recovery state
in which the thrusters are recovered. 
$U2$
influences the maximum distance at which the AUV can visually perceive
objects. This is relevant for $T1$, higher water visibility
allows the AUV to search for the pipeline at higher altitudes,
resulting in a larger field of view and the possibility of discovering the pipeline faster. On the other hand, if the water
visibility is low, the AUV has to move closer to the seabed to
search for the pipeline, which limits its field of view and therefore
may lead to a longer time to discover the pipeline. Thus, changing the
altitude of the AUV provides functional alternatives for dealing with $U2$.

This exemplar focuses on the problem of overcoming $U1$ and $U2$ using
a self-adaptation logic, implemented by a managing subsystem, that
can be extended and reused for other sources of uncertainty.
The managing subsystem shall overcome $U1$ by recovering the failed
thrusters at runtime, and $U2$ by adapting the maximum altitude for
the path generator algorithm according to the measured water
visibility.  Thus, by reacting to $U1$ and $U2$, the managing
subsystem increases the reliability and performance of the system.

For the feasibility of the exemplar, the use case was simplified while still allowing for a worthwhile application of self-adaptation to an AUV.  
 It is important to highlight that a
realistic operation of an AUV used for pipeline inspection would include
steps that are related to pre-dive, launching and recovery, human
interaction, and intermediary missions that are necessary to enable
the inspection.  Furthermore, there are several sources of uncertainty
not considered here, including ocean dynamics, sensor failures, and
battery duration.

\subsection{System Architecture}

\begin{figure}[h]
    \centering
    \includegraphics[width=0.8\linewidth]{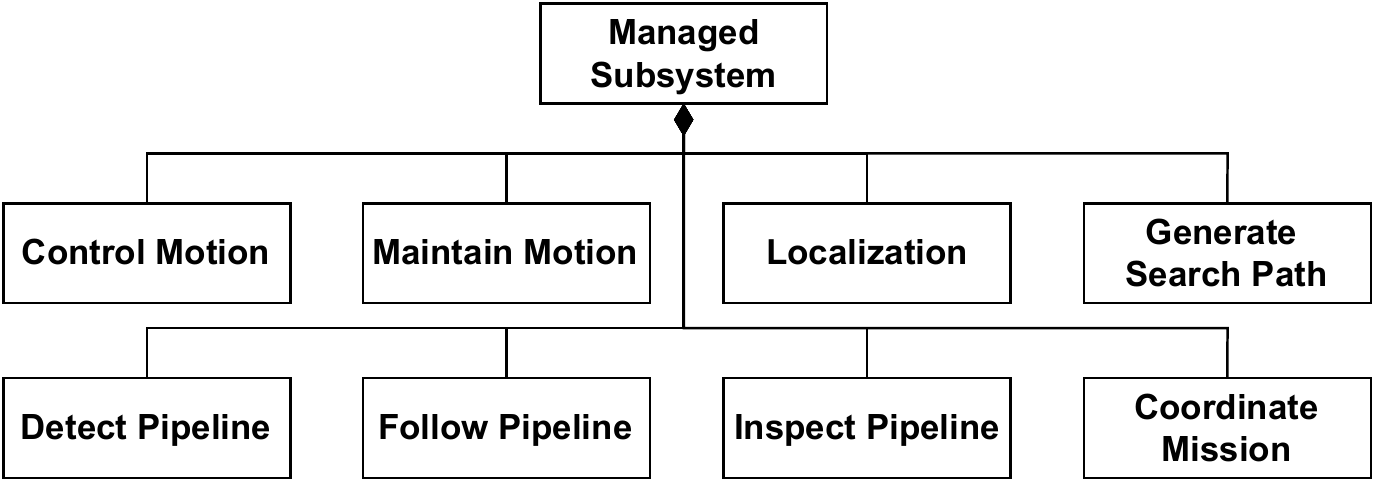}
    \caption{Managed Subsystem's Functional Hierarchy}
    \label{fig:func_arch}
\end{figure}

To accomplish the mission described in Section~\ref{sec:use_case_description}, the managed subsystem requires the
functions represented in Fig.~\ref{fig:func_arch}. $T1$ requires the
functions \kw{Control Motion}, \kw{Maintain Motion},
\kw{Localization}, \kw{Detect Pipeline}, \kw{Generate Search
  Path}, and \kw{Coordinate Mission}, while $T2$ requires the functions \kw{Control Motion},
\kw{Maintain Motion}, \kw{Localization}, \kw{Detect Pipeline},
\kw{Follow Pipeline}, \kw{Inspect Pipeline}, and \kw{Coordinate Mission}. During runtime, the
functions must be activated and deactivated according to the task
being performed.

To overcome the uncertainties $U1$ and $U2$, a managing subsystem
requires the functionalities to \emph{monitor} the environment and the
managed subsystem's internal state, \emph{reason} about it, and
\emph{execute} the managed subsystem's reconfiguration.

\begin{figure}
    \centering
    \includegraphics[width=\linewidth]{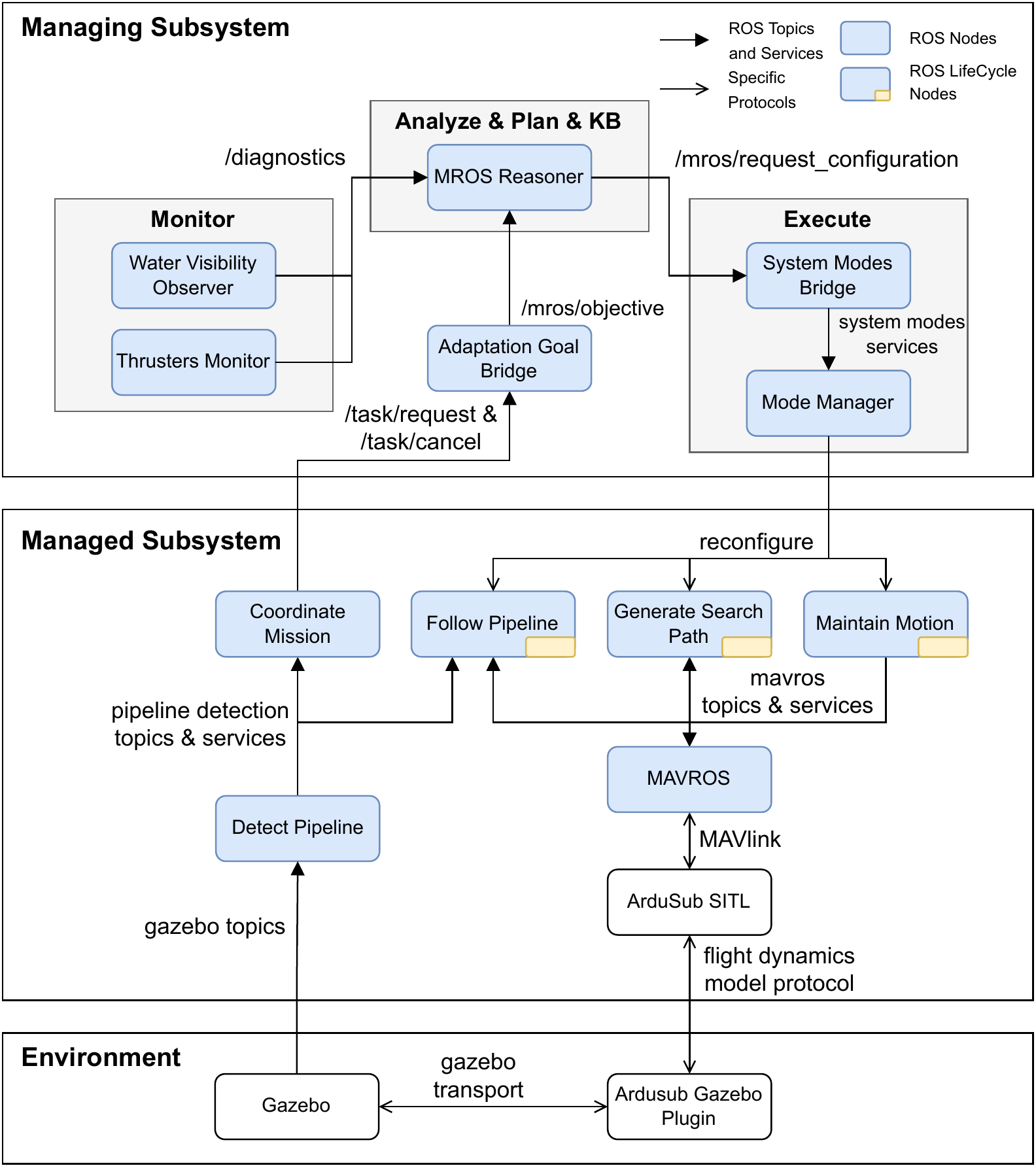}
    \caption{System Architecture}
    \label{fig:system_architecture}
\end{figure}

The required functions of the managed and managing subsystems
are realized as depicted in Fig.~\ref{fig:system_architecture}. The
managed subsystem is detailed in Section~\ref{sec:managed} and the
managing subsystem in Section~\ref{sec:managing_subsystem}.
It is important to mention that managed subsystem functions
\kw{Control Motion} and \kw{Localization} are achieved by \kw{ArduSub}, and the function \kw{Inspect Pipeline} is not realized since
the actual inspection of the pipeline is not the focus of this
work. It is also important to highlight that this exemplar implements
the function to \emph{reason} about the managing subsystem with Metacontrol to provide a
baseline for future research. However, it can be replaced with other
solutions, as long as they are compatible with the monitor and execute
interfaces, as described in Section~\ref{sec:managing_requirements}.

\section{Managed Subsystem} \label{sec:managed}

The managed subsystem is implemented as a ROS2-based system and is depicted in Fig.~\ref{fig:system_architecture}. The only non-ROS2 component is \kw{ArduSub}\footnote{\url{https://www.ardusub.com/}}, which is an open-source autopilot for underwater vehicles. In this application it is used to solve the functions \kw{Control Motion} and \kw{Localization}\footnote{It is assumed that the AUV has appropriate sensors for localization}. The \kw{MAVROS}
package works as a bridge between \kw{ArduSub} and the ROS2 components. The \kw{Detect Pipeline} node detects the pipeline and informs \kw{Follow Pipeline} and the \kw{Coordinate Mission} node about its position\footnote{A mock perception system is used.}. The \kw{Coordinate Mission} node coordinates the tasks' execution and sets the adaptation goals. Note that the function \kw{Inspect Pipeline} is not implemented,
since the actual inspection of the pipeline is not the focus of this work. However, the exemplar can easily be extended with this functionality by adding a new node that implements the pipeline inspection.

\kw{Follow Pipeline}, \kw{Generate Search Path},  and \kw{Maintain Motion} are lifecycle nodes, which means that they have internal states, such as \emph{active} and \emph{inactive}, and it is possible to switch between these states at runtime. Furthermore, the System Modes package \cite{nordmann2021system} 
extends the state \emph{active} with additional modes, e.g., \emph{active.low\_altitude}.

To adapt the managed subsystem, the managing subsystem adapts the lifecycle nodes by changing their states. This is done by the \kw{Mode Manager} node, which is used off-the-shelf from the System Modes package. 
The states available for \kw{Generate Search Path} are \kw{deactivated}, \kw{low altitude}, \kw{medium altitude}, and \kw{high altitude}. Subsequently, the states available for \kw{Follow Pipeline} are \kw{deactivated} and \kw{activated}, while the states for \kw{Maintain Motion} are \kw{deactivated}, and \kw{recover thrusters}.

To enable other developers of self-adaptive systems to use this exemplar and compare different approaches, a \kw{Gazebo}-based  \footnote{\url{https://gazebosim.org/home}} simulation of a pipeline inspection environment and a model of the AUV is provided. The BlueROV2\footnote{\url{https://bluerobotics.com/store/rov/bluerov2/}}
robot was selected as the AUV for the exemplar because (i)~it is compatible with \kw{ArduSub}; (ii)~it is easily integrated with \kw{Gazebo} via plugins; and (iii)~the robot has a low price compared to other available AUVs, making it more accessible to researchers to reproduce the exemplar with a real robot.

\section{Managing Subsystem}
\label{sec:managing_subsystem}

The managing subsystem exploits functional alternatives of the managed
subsystem to enable adaptation and thereby increase system reliability.
Metacontrol is used as an example of how a managing
subsystem can be implemented. This section introduces
Metacontrol
and shows how the adaptation problem can be
formulated and implemented with Metacontrol.

\subsection{Metacontrol Background}
\label{subsec:metacontrol}
\emph{Metacontrol} uses the MAPE-K feedback loop \cite{brun09,kephart03computer} to implement self-adaptation. It \emph{Monitors} the managed subsystem during runtime, \emph{Analyzes} whether the system meets its requirements, \emph{Plans} a new configuration if the system does not meet the requirements, and then \emph{Executes} the reconfiguration of the managed subsystem. All this is done using a shared \emph{Knowledge Base} to which each step refers. In Metacontrol, the knowledge base conforms to the TOMASys (Teleological and Ontological Metamodel for Autonomous Systems) metamodel \cite{corbato2013model}. 

\begin{figure}
    \centering
    \includegraphics[width=0.4\textwidth]{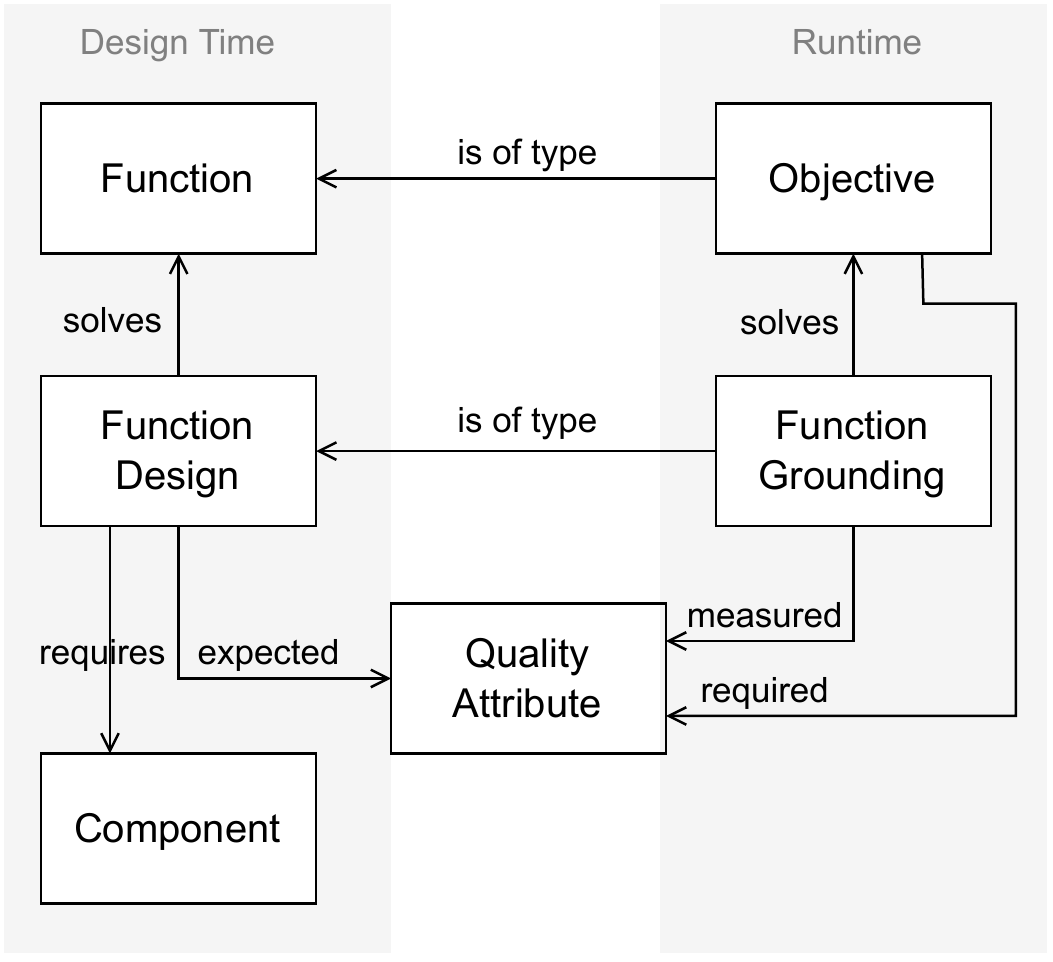}
    \caption{A simplified representation of the TOMASys elements}
    \label{fig:tomasys}
\end{figure}

A simplified version of the TOMASys metamodel is displayed in Fig.~\ref{fig:tomasys}.
TOMASys uses \emph{functions} $F$ to represent the functionalities of the system, e.g., generating a search path for the AUV.
The architectural variants that implement these requirements are captured by \emph{function designs} $FD(F, \mathcal{C}, \qa^{exp})$. To distinguish during runtime which function design is most suited in a given situation, a set $\qa^{exp}$ of \emph{expected quality attributes} is associated with it. An expected quality attribute value reflects how well a function design is supposed to fulfill the function $F$ it solves. Furthermore, a function design requires a set $\mathcal{C}$ of \emph{components} of the managed subsystem to solve $F$. A component $C(S_C)$ is a piece of hardware or software, e.g., a sensor or a path-planning algorithm, respectively. The status $S_C$ of a component indicates its availability, i.e., whether it is functioning or not.

An \emph{objective} $O(F,S_O,\qa^{req})$ is a runtime instantiation of a function $F$, e.g., generating a search path with a minimum required water visibility, whose status $S_O$ reflects whether the objective is currently achieved. Furthermore, the set $\qa^{req}$ of \emph{required quality attributes} specifies which quality attribute value the objective requires in order to work properly.

An objective $O$ is solved by a \emph{function grounding} $FG(O,FD,S_{FG}, \qa^{meas})$, which represents the function design $FD$ that is currently used to solve the objective. 
Its status $S_{FG}$ reflects whether the function grounding is currently able to achieve the objective. The set $\qa^{meas}$ of \emph{measured quality attributes} reflects how well the function grounding currently fulfills $O$ and is computed using sensor data.

\subsection{Metacontrol Formulation}
\label{sec:metacontrol_formulation}
The functions, architectural variants, and quality attributes required to solve the tasks $(T1)$ Search Pipeline and $(T2)$ Inspect Pipeline, described in Section \ref{sec:problem_statement}, are modeled conforming to the TOMASys metamodel. Table~\ref{tbl:functions_exemplar} specifies the functions ($F_1$)~maintain\_motion, ($F_2$)~generate\_search\_path, and ($F_3$)~follow\_pipeline, while Table~\ref{tbl:qas_exemplar} describes the quality attributes ($QA_1$)~water\_visibility, and ($QA_2$)~performance. 
Functions $F_1$ and $F_2$ are required to achieve $T1$, whereas $T2$ is achieved by $F_1$ and $F_3$.
The function designs that solve these functions are specified in Table~\ref{tbl:function_designs_exemplar}. The set of required components is empty for function designs $FD_2 - FD_6$ because they do not require any components that are susceptible to adaptation, or used in the reasoning process.

Since the objectives and function groundings are instantiated during runtime, they are not specified here. An objective for function $F_2$ is for example to generate a search path with no required quality attribute, which is defined as $O_2(F_2, \text{\emph{ok}}, \varnothing)$ in the notation introduced above. A possible function grounding for this objective is $FG_2(O_2, FD_4, \text{\emph{ok}}, \{QA_1^{meas}=1.1\})$.

\begin{table}
    \centering
    \caption{The TOMASys functions used for the exemplar}
    \label{tbl:functions_exemplar}
    \begin{tabular}{|@{\;}l@{\;}|@{\;}l@{\:}|p{4.7cm}@{\:}|}
        \hline
        \textbf{Function} & \textbf{Name} & \textbf{Requirement} \\
        \hline
        \rule{0pt}{2ex} 
        $F_1$ & maintain\_motion & Maintain the motion of the robot\\
        \hline
        \rule{0pt}{2ex} 
        $F_2$ & generate\_search\_path & Generate a path to search for the pipeline\\
        \hline
        \rule{0pt}{2ex} 
        $F_3$ & follow\_pipeline & Follow and inspect the pipeline\\
        \hline
    \end{tabular}
\bigskip
\caption{The TOMASys quality attributes used for the exemplar}
    \label{tbl:qas_exemplar}
    \begin{tabular}{|l|l|l|p{3.9cm}|}
        \hline
        \rule{0pt}{2ex} 
        \textbf{QA} & \textbf{Name} & \textbf{Unit} & \textbf{Description}\\
        \hline
        \rule{0pt}{2ex} 
        $QA_1$ & water\_visibility & [$0,\infty)$ & Reflects the maximum altitude (in meters) from which the AUV can perceive the seabed\\
        \hline
        \rule{0pt}{2ex} 
        $QA_2$ & performance & $[0,1]$ & Reflects how efficient the current search strategy is\\
        \hline
    \end{tabular}
\bigskip
\caption{The TOMASys function designs used for the exemplar}
    \label{tbl:function_designs_exemplar}
    \begin{tabular}{|@{\;}p{3.5cm}@{\;\,}|@{\;}p{2.0cm}@{\,}|p{2.7cm}@{\;}|}
        \hline
        \textbf{Function Design} & \textbf{Name} & \textbf{Description} \\
        \hline
        \rule{-3pt}{2ex} 
        $FD_1(F_1,\{\text{thruster}\_x\mid x=1,\dots,6\}, \{QA_2^{exp}=1\})$ & all\_thrusters & Uses all thrusters \\
        \hline
        \rule{-3pt}{2ex} 
        $FD_2(F_1, \varnothing, \{QA_2^{exp}=0.5)$ & recover\_thrusters & Recovers the thrusters that are in failure\\
        \hline
        \rule{-3pt}{2ex} 
        $FD_3(F_2, \varnothing, \{QA_1^{exp}=0.5,$ $ QA_2^{exp}=0.25\})$ & spiral\_low & Produces a spiral search path with low altitude \\
        \hline
        \rule{-3pt}{2ex} 
        $FD_4(F_2, \varnothing, \{QA_1^{exp}=1,$ $ QA_2^{exp}=0.5\})$ & spiral\_medium & Produces a spiral search path with medium altitude \\
        \hline
        \rule{-3pt}{2ex} 
        $FD_5(F_2, \varnothing, \{QA_1^{exp}=2,$ $ QA_2^{exp}=1\})$ & spiral\_high & Produces a spiral search path with high altitude \\
        \hline
        \rule{-3pt}{2ex} 
        $FD_6(F_3, \varnothing, \varnothing)$ & follow\_pipeline & Follows the pipeline \\
        \hline
    \end{tabular}
\end{table}

The MAPE-K loop steps in this exemplar are formulated as follows. The monitor step is responsible for measuring $QA_1^{meas}$ and for monitoring the state of the six thrusters. 

The analyze step uses Horn rules
to reason about the knowledge base. One example rule that analyzes whether the measured water visibility $QA_1^{meas}$ still satisfies the expected water visibility $QA_1^{exp}$ of the grounded function design is displayed in Fig.~\ref{fig:swrl_rule}. Note that it is written in terms of the notation introduced in Section~\ref{subsec:metacontrol}. Line~\ref{eq:fg} expresses that the rule reasons about a function grounding $FG$ that solves an objective $O$, is of type $FD$, has a status $S_{FG}$ and an associated set of measured quality attributes $\qa^{meas}$. Furthermore, the function design $FD$ solves the function $F$, has a set of required components $\mathcal{C}$ and an associated set of expected quality attributes $\qa^{exp}$. Note that it is implicitly assumed that $FG$ is well-formed, i.e., that the function of which $O$ is a type of is the same as the function that $FD$ solves. Since this rule should analyze the water visibility, Line~\ref{eq:qas_contained} ensures that $QA_1^{meas}$ is an element of the set $\qa^{meas}$ and that $QA_1^{exp}$ is an element of the set $\qa^{exp}$, i.e., that both $FG$ and $FD$ are related to water visibility. Finally, if the measured value of $QA_1$ is less than its expected value associated with the grounded function design, see Line \ref{eq:meas_less_exp}, then the status of the function grounding is set to \emph{error}, see Line~\ref{eq:status_error}.

\begin{figure}[t]
    \centering
    \begin{align}
    &FG\bigl(O, FD, S_{FG}, \qa^{meas}\bigr) \boldsymbol{\wedge} ~ FD(F, \mathcal{C}, \qa^{exp}) \label{eq:fg}\\
    & \boldsymbol{\wedge} ~ QA_1^{meas} \in \qa^{meas}
    ~\boldsymbol{\wedge} ~ QA_1^{exp} \in \qa^{exp} \label{eq:qas_contained}\\
    &\boldsymbol{\wedge} ~ QA_1^{meas} < QA_1^{exp} \label{eq:meas_less_exp}\\
    &\boldsymbol{\Rightarrow} ~ S_{FG} = \text{\emph{error}} \label{eq:status_error}
\end{align}
\caption{Rule to analyze whether the measured water visibility $QA_1^{meas}$ still satisfies the expected water visibility $QA_1^{exp}$ of the grounded function design}
\label{fig:swrl_rule}
\end{figure}

In the planning step, the function designs with $QA_1^{exp}$ higher than $QA_1^{meas}$ are filtered out as the visibility they would expect is not measured, afterward the remaining function design with the highest expected search performance ($QA_2^{exp}$) is selected as the desired configuration. The selected configuration is then carried out in the execute step.

\subsection{Metacontrol Implementation}

As depicted in Fig.~\ref{fig:system_architecture}, the monitor step is implemented with the \kw{Water Visibility Observer} and the \kw{Thruster Monitor} nodes. They are used for measuring $QA_1$ and monitoring the status of the six thrusters $\text{thruster}\_x$ where $x\in\{1,\dots,6\}$, respectively. 
To simplify the system and avoid the addition of unnecessary nodes, instead of adding water visibility to the Gazebo simulator, the \kw{Water Visibility Observer} simulates water visibility measurements with a sine function, and instead of probing the managed subsystem to identify thruster failures the \kw{Thruster Monitor} simulates the thruster failures events. Since the monitor step is mocked up, and its probes and intermediary nodes that would be required to provide the probes are not implemented, they are not included in Fig. \ref{fig:system_architecture}. Both nodes publish their data into the \kw{/diagnostics} topic with the ROS2 default \kw{DiagnosticArray} message type.

The knowledge base (KB), the analyze and plan step are implemented using MROS2~\footnote{\url{https://github.com/meta-control/mc_mros_reasoner}}~\cite{corbatoMROSRuntimeAdaptation2020}, a ROS2-based Metacontrol implementation, as the \kw{MROS Reasoner} node. The KB
is implemented with the Ontology Web Language (OWL)~\cite{Antoniou2004}, the Horn rules used for the analyze step with the Semantic Web Rule Language (SWRL)~\cite{horrocks2004swrl}, and the reasoning is done with Pellet\footnote{\url{https://github.com/stardog-union/pellet}}. The \kw{MROS Reasoner} receives water visibility measurements ($QA_1^{meas}$) and thruster status information from the monitor step, then decides whether adaptation is required, and, in this case, selects a desired configuration which it sends to the execute step (see Section \ref{sec:metacontrol_formulation} for more details). The \kw{MROS Reasoner} initially does not have objectives, so it does not perform adaptation. The adaptation reasoning only starts when the \kw{Coordinate Mission} node sends new objectives, such as $O_2(F_2, null , \varnothing)$, via the \kw{Adaptation Goal Bridge}. New objectives do not have a status yet.

The execute step uses the System Modes' \kw{Mode Manager} to adapt the managed subsystem, and the \kw{System Modes Bridge} bridges the \kw{Mode Manager} with the \kw{MROS Reasoner}. When a reconfiguration is needed, the \kw{MROS Reasoner} requests the new configuration via the \kw{/mros/request\_configuration} service to the \kw{System Modes Bridge}.Then the \kw{System Modes Bridge} forwards the request to the \kw{Mode Manager} using the correct service names, depending on the lifecycle node being adapted. The services used by the \kw{Mode Manager} are listed in Table~\ref{tbl:system_modes_services}, and the available modes are listed in Table~\ref{tbl:system_modes_modes}.

\begin{table}
    \centering
    \caption{Available System Modes' services}
    \label{tbl:system_modes_services}
    \begin{tabular}{|p{3cm}|p{4.4cm}|}
        \hline
        \textbf{Node} & \textbf{Service} \\
        \hline
        f\_generate\_search\_path  & /f\_generate\_search\_path/change\_mode   \\
        \hline
        f\_follow\_pipeline  & /f\_follow\_pipeline/change\_mode   \\
        \hline
        f\_maintain\_motion  & /f\_maintain\_motion/change\_mode   \\
        \hline
    \end{tabular}
    \bigskip
    \caption{Available modes}
    \label{tbl:system_modes_modes}
    \begin{tabular}{|p{3cm}|p{2.5cm}|c|}
        \hline
        \textbf{Node} & \textbf{Mode} & \textbf{Lifecycle state} \\
        \hline
        f\_generate\_search\_path & fd\_spiral\_high   & active               \\
        \hline
        f\_generate\_search\_path & fd\_spiral\_medium   & active             \\
        \hline
        f\_generate\_search\_path & fd\_spiral\_low   & active              \\
        \hline
        f\_generate\_search\_path & fd\_unground   & inactive  \\
        \hline
        f\_follow\_pipeline & fd\_follow\_pipeline   & active  \\
        \hline
        f\_follow\_pipeline & fd\_unground   & inactive  \\
        \hline
        f\_maintain\_motion & fd\_all\_thrusters   & inactive  \\
        \hline
        f\_maintain\_motion & fd\_recover\_thrusters   & active  \\
        \hline
    \end{tabular}
\end{table}

\section{Extending and connecting managing subsystems}\label{sec:managing_requirements}

With the described system implementation, the only Metacontrol-specific nodes of the system are the \kw{MROS Reasoner}, the \kw{System Modes Bridge}, and the \kw{Adaptation Goal Bridge}. All other nodes of the system can be reused with different managing subsystems. The only requirement to connect a managing subsystem to the managed subsystem is to ensure that the managing subsystem adheres to the provided monitor and execute ROS2 interfaces. As described in the previous section, the monitor interface is the ROS2 topic \kw{/diagnostics}, and the execute interfaces are listed in Table~\ref{tbl:system_modes_services}. To show that changing the managing subsystem is possible, a managing subsystem that randomly picks a configuration was also implemented.

Since the system is implemented with a modular design, it can be extended with additional functionalities and adaptation scenarios by adding new lifecycle nodes and updating the system modes' configuration file accordingly. The implemented functionalities can be replaced with different implementations as long as they adhere to the same interfaces; e.g., the \kw{Pipeline Detection} node could be replaced by a node that actually performs perception instead of a mock-up.

\section{Evaluation}\label{sec:results}
To evaluate the performance of different managing subsystems using this exemplar, the mission described in Section~\ref{sec:problem_statement} was implemented.
The mission consists of the AUV performing $T1$ and $T2$ while subject to $U1$ and $U2$ until a user-provided time limit  is reached. To evaluate the mission, the following metrics were used: the \emph{search time}, the amount of time elapsed from the beginning of the search until the pipeline is found, and the total \emph{distance inspected} of the pipeline.

To provide a baseline for the exemplar, the mission is performed with two different managing subsystems and with no managing subsystem, using a fixed configuration. The managing subsystems are the Metacontrol-based implementation detailed in Section~\ref{sec:managing_subsystem} and a random managing subsystem that selects configurations arbitrarily. 

Since the system is non-deterministic due to characteristics of \kw{Gazebo}, \kw{ArduSub}, and the interaction between them, no run of the simulation is exactly the same. Thus, the mission execution and metrics collection are automated with a runner to allow multiple runs to be easily performed. 

This section briefly describes how to configure the exemplar, and the results of running the exemplar. Further details may be found in the exemplar repository.

\subsection{Configuring the exemplar}

In \acronym, the AUV's mission execution can be varied by changing the parameters of the system. In the \kw{Water Visibility Observer}, the available parameters are the water visibility minimum and maximum values, periodicity, and initial phase shift. In the \kw{Thrusters Monitor}, the available parameter is a list with thruster events indicating which thruster fails and when. In the \kw{Coordinate Mission}, the mission time limit can be set. In the random manager, the adaptation periodicity can be set, and when using no manager, the default states for the lifecycle nodes can be set. In addition, the runner is parametrized with the number of runs to execute, and which managing subsystem to select. All parameters are adjusted using configuration files packaged in the exemplar.

\subsection{Results}
The mission was executed with a time limit of 300 seconds, water visibility  periodicity of 80 seconds, minimum and maximum values of 1.25 and 3.75, no phase shift, and thruster 1 failing after 35 seconds from the start of the mission. The results are shown in Table~\ref{tab:results_m1}. It can be noticed that with the Metacontrol managing subsystem both mean \emph{search time}\footnote{When the pipeline is not found, the time limit is used as the \emph{search time}} is lower, and the \emph{distance inspected} is higher. 
This indicates that, in this exemplar, Metacontrol improves the performance of the system, and outperforms the random managing subsystem and the system without a managing subsystem. 
In addition, the standard deviation (Std) of the \emph{search time} is lower for Metacontrol, indicating that it is more consistent when searching for the pipeline. The Std of the random manager for the \emph{distance inspected} is lower, however, its mean value is also low, indicating that the random manager is consistent in not inspecting the pipeline.
The results shown can be used as a baseline for comparing different managing subsystems.

\begin{table}[t]
\caption{Mission results}
\label{tab:results_m1}
\resizebox{\columnwidth}{!}{%
\begin{tabular}{|l|l|ll|ll|}
\hline
\multirow{2}{*}{\textbf{\begin{tabular}[c]{@{}l@{}}Managing\\ subsystem\end{tabular}}} &
  \multirow{2}{*}{\textbf{\begin{tabular}[c]{@{}l@{}}Number \\ of runs\end{tabular}}} &
  \multicolumn{2}{l|}{\textbf{Search time (s)}} &
  \multicolumn{2}{l|}{\textbf{Distance inspected (m)}} \\ \cline{3-6} 
                     &            & \multicolumn{1}{l|}{\textbf{Mean}} & \textbf{Std} & \multicolumn{1}{l|}{\textbf{Mean}} & \textbf{Std} \\ \hline
None                 & 20          & \multicolumn{1}{l|}{187.85}             & 42.40            & \multicolumn{1}{l|}{31.40}             & 13.00            \\ \hline
Random               & 20          & \multicolumn{1}{l|}{180.15}             & 82.05           & \multicolumn{1}{l|}{3.88}             & 3.19            \\ \hline
\textbf{Metacontrol} & \textbf{20} & \multicolumn{1}{l|}{\textbf{106.95}}    & \textbf{39.46}   & \multicolumn{1}{l|}{\textbf{51.15}}    & \textbf{12.01}   \\ \hline
\end{tabular}%
}
\end{table}

\section{Conclusion}\label{sec:conclusion}

This work describes \acronym, a ROS2-based exemplar for self-adaptive underwater vehicles used for pipeline inspection. Due to its modular design, \acronym enables different managing subsystems to be applied to the system without the need to modify the managed subsystem, the monitor nodes, and the executing mechanism. In addition, the system can be easily extended with new functionalities and adaptation scenarios by adding new nodes. Furthermore, this paper provides a baseline for comparing the performance of different managing subsystems, and it shows that the addition of a Metacontrol-based managing subsystem increases the performance of the system in comparison to not using any managing subsystem or one that chooses configurations arbitrary.

In future work, \acronym can be extended with: more metrics for a more in-depth evaluation; more tasks (e.g., docking), functionalities (e.g. a de facto perception system), and components (e.g. sonars) for more realistic missions; more adaptation scenarios, e.g., adapting to changes in the water currents, and adapting the thruster configuration matrix when a thruster can not be recovered.

\end{document}